\documentclass[final,5p,times,twocolumn]{elsarticle}

\usepackage{amssymb}

\usepackage{graphicx}
\usepackage{amsmath}
\usepackage{amssymb}
\usepackage{booktabs}
\usepackage{float}
\usepackage{multirow}
\usepackage{url}
\usepackage[colorlinks,
            linkcolor=red,
            anchorcolor=blue,
            citecolor=green
            ]{hyperref}
\usepackage{stfloats}
\usepackage{makecell, tabularx}
\usepackage[switch]{lineno}

\journal{Journal}

\begin{document}

\begin{frontmatter}



\title{RFR-WWANet: Weighted Window Attention-Based Recovery Feature Resolution Network for Unsupervised Image Registration}


\author[label1,label2]{Mingrui Ma}
\ead{mamr@mails.jlu.edu.cn}

\author[label1,label2]{Tao Wang}
\ead{taowang19@mails.jlu.edu.cn}

\author[label3]{Weijie Wang}
\ead{weijie.wang@unitn.it}

\author[label1,label2]{Lei Song}
\ead{songlei20@mails.jlu.edu.cn}

\author[label1,label2]{Guixia Liu\corref{mycorrespondingauthor}}
\ead{liugx@jlu.edu.cn}
\cortext[mycorrespondingauthor]{Corresponding author}

\address[label1]{College of Computer Science and Technology, Jilin University, Changchun, China}

\address[label2]{Key Laboratory of Symbolic Computation and Knowledge Engineering of Ministry of Education, Changchun, China}

\address[label3]{Department of Information Engineering and Computer Science, University of Trento, Trento, Italy}

\begin{abstract}

The Swin transformer has recently attracted attention in medical image analysis due to its computational efficiency and long-range modeling capability. Owing to these properties, the Swin Transformer is suitable for establishing more distant relationships between corresponding voxels in different positions in complex abdominal image registration tasks. However, the registration models based on transformers combine multiple voxels into a single semantic token. This merging process limits the transformers to model and generate coarse-grained spatial information. To address this issue, we propose Recovery Feature Resolution Network (RFRNet), which allows the transformer to contribute fine-grained spatial information and rich semantic correspondences to higher resolution levels. Furthermore, shifted window partitioning operations are inflexible, indicating that they cannot perceive the semantic information over uncertain distances and automatically bridge the global connections between windows. Therefore, we present a Weighted Window Attention (WWA) to build global interactions between windows automatically. It is implemented after the regular and cyclic shift window partitioning operations within the Swin transformer block. The proposed unsupervised deformable image registration model, named RFR-WWANet, detects the long-range correlations, and facilitates meaningful semantic relevance of anatomical structures. Qualitative and quantitative results show that RFR-WWANet achieves significant improvements over the current state-of-the-art methods. Ablation experiments demonstrate the effectiveness of the RFRNet and WWA designs. Our code is available at \url{https://github.com/MingR-Ma/RFR-WWANet}.

\end{abstract}




\end{frontmatter}


\section{Introdcution}

In the past decade, Convolutional Neural Networks (CNNs) have achieved significant success in computer vision (CV). Benefiting from these successes and the rapid development of CNNs, CNN-based approaches \cite{2015-Unet,nnunet,2018-VM} became a significant focus in medical image analysis. Especially since the U-Net \cite{2015-Unet} was proposed, due to its ability to effectively incorporate both low-level and high-level semantic information with a limited number of parameters, it and its variants \cite{2018-unet++-zhou,2019-resunet++} have been widely utilized in medical image analysis tasks. For image registration, a fundamental study in medical image analysis, CNN-based approaches \cite{pr1,pr2} have also become a hot research topic in recent years. CNN-based methods \cite{2018-VM,2019-PVM,2020-SYM,2021-CycleMorph} have the advantages of registration accuracy and fast prediction compared with traditional methods, including \cite{LDDMM,deeds,SyN}. However, due to the limited receptive field range of CNNs, the registration performance may be restricted \cite{2020-CNNLimitation-VIT,2016-CNNLimitation,transmorph}.

In recent years, transformer-based methods \cite{2017-attention,BERT,alBERT} have achieved remarkable achievements in natural language processing (NLP) due to the self-attention mechanism, which models tokens on a global scale. Dosovitskiy et al. \cite{2020-CNNLimitation-VIT} introduce the transformer into the field of CV and achieves promising results in image recognition, which makes people realize the potential of the transformer and attracts attention to widely utilize it in the field of CV. Deformable image registration sensitive to spatial correspondences within uncertain ranges may also be suitable to be modeled by transformers.

Liu et al. propose Shifted window transformer (Swin transformer) \cite{Swintrans}, a hierarchical transformer-based architecture, which performs computing of multi-head self-attention (MSA) within each window by window partitioning. In this way, the complexity of MSA in the Swin transformer is reduced from quadratic complexity to linear complexity, which means the Swin transformer is more efficient than the standard transformer. Furthermore, the hierarchical nature of the Swin transformer makes it more suitable for multi-scale modeling tasks. Recently, the Swin transformer-based TransMorph \cite{transmorph} is proposed, and its results demonstrate the outperformance in deformable image registration.

Abdominal image registration is challenging due to the complex anatomical structures present in the abdomen, which can vary significantly in size, shape, and position between individuals. This makes it difficult to establish correspondences between voxels in different images, particularly in cases where there are significant variations or deformations, such as in patients with tumors or other pathologies. One of the main difficulties in abdominal image registration is accurately aligning structures that have different shapes and sizes, such as the liver, spleen, and pancreas, or the deformations caused by respiratory motion or organ displacement, which can lead to significant misalignments between abdominal images \cite{organ_sliding}.
The success of long-range modeling is illustrated by recent studies on transformers, which demonstrate the potential of transformers to establish distant voxel correlations in abdominal image registration tasks.

However, two problems exist with using Swin Transformer directly in the abdominal registration model:

\textbf{(i)} \textbf{Lack of fine-grained spatial information.} Due to a large number of parameters of the Transformer, generally, $4\times4\times4$ voxels are input into it as a token when using the Transformer \cite{transmorph,my,2021-swinunet,SymSwin}, and the Transformer can only output information of the same scale. Thus, the output representations of the transformer lose the fine-grained spatial information essential for dense deformation field prediction.

\textbf{(ii)} \textbf{Inflexible window connection.} The Swin transformer uses the shifted window partition operation to establish the connection between windows to address the limitation of the modeling ability of each regular partitioned window. For abdominal images, due to the sliding of human organs and the inconsistency of human postures, it is difficult to match the same anatomical structure in a pair of images following the same coordinate system and the same anatomical structure may also be far apart. Therefore, global information interaction after the window partitioning operation may improve the modeling performance of the abdominal registration model.



To address these challenges, we introduce the \textbf{R}ecovery \textbf{F}eature \textbf{R}esolution \textbf{Net}work (RFRNet), a U-shaped model based on the Swin Transformer. The RFRNet consists of a Swin Transformer-based encoder and a CNN-based decoder. The encoder captures the feature representations of an image pair while the decoder restores these representations to form a dense deformation field. After the first two Swin Transformer blocks, the feature representation recovery blocks are utilized to channel-wise supplement the representations and restore the resolution to higher stages, then connect the recovered feature representations to the decoder using the skip connections. This process supplements the spatial information and enhances the contribution of the representations output from the first two Swin transformer blocks to higher-resolution stages.

Furthermore, we propose the \textbf{W}eighted \textbf{W}indow \textbf{A}ttention (WWA) mechanism for the partitioned windows. The WWA mechanism creates connections among the windows by learning the global information of each window and determining how to adjust their feature representations. This approach dynamically and automatically establishes connections between windows. The proposed network and attention mechanism in this paper are combined and referred to as the RFR-WWANet.

In summary, the contributions of this work are as follows:
\begin{itemize}
   \item We propose an architecture for unsupervised image registration called RFRNet, which consists of a Swin transformer-based encoder and a CNN-based decoder. RFRNet augments the contributions of the first two Swin Transformer blocks by restoring feature representations to higher-resolution stages in the decoder while enabling the output of fine-grained spatial information.
   \item We propose a weighted window attention mechanism, dubbed WWA, to automatically establish the connections between windows to achieve global information interactions.
   \item We validate our proposed model RFR-WWANet on the 3D abdominal datasets, and the experimental results demonstrate the state-of-the-art performance of the proposed method. And the ablation studies illustrate that our RFRNet and WWA are effective.
\end{itemize}

\section{Related Work}
Deformable image registration aims to establish spatial correspondence between an image pair. Current deformable image registration approaches can be divided into traditional and deep learning-based methods.

\subsection{Traditional Approaches}
Deformable image registration models have been improved rapidly over the past decades. Traditional deformable image registration methods iteratively optimize the similarity functions to find the optimal deformation field. The conventional image registration methods, such as LDDMM \cite{LDDMM}, SyN \cite{SyN}, Demons \cite{demons}, and deeds \cite{deeds}, face the problem of time-consuming calculations.

\subsection{Deep Learning-Based Approaches}
Image registration methods based on CNNs extract deep feature representations of image pairs and utilize similarity loss functions to train models. The CNN-based methods predict the deformation field of an image pair in a short time after training, and the CNN-based methods \cite{dir,2018-VM,2011-diffeomorphic,2020-diff-DeepFlash,2019-Cascade,2020-SYM} have demonstrated superior performance than traditional approaches. Since the ground-true deformation fields are difficult to obtain, this limits supervised learning methods \cite{2017-supervisedRegistration-1,2017-supervisedRegistration-2,supervised_18} in practical application.

Unlike supervised approaches, unsupervised CNN-based approaches do not require ground-truth information. The unsupervised deformable image registration approaches \cite{2018-VM,2019-PVM,unsupervised_2018_ICCV,unsupervised_2018_MIA,unsupervised_2020_CVPR} have been brought to the fore. The unsupervised method, such as VoxelMorph \cite{2018-VM}, introduces a U-shaped registration framework, which predicts the dense displacement vector field. Dalca et al. present the diffeomorphic registration model \cite{2019-PVM} that utilizes the stationary velocity fields to achieve topology preservation in registration. Mok et al. \cite{2020-SYM} propose a symmetric model which guarantees invertibility and diffeomorphic property. Kim et al.\cite{2021-CycleMorph} use cycle consistency to enhance the registration performance and preserve topology. Although CNN-based methods have achieved great success, their performance is still limited by the shortcomings of CNNs, i.e., the limited receptive field of CNNs.

Since vision transformers (ViTs) developed rapidly, some latest ViT-based studies based on have been proposed. Chen et al.\cite{2021-vitV} introduce ViT into V-Net. Zhang et al. \cite{2021-zhang-DTN} propose a dual ViT-based network to enhance the feature modeling. Ma et al. \cite{my} present a symmetric variant ViT-based U-Net to improve the registration performance. \cite{transmorph} proposes TransMoprh, consisting of a Swin transformer-based encoder and a CNN-based decoder. \cite{SymSwin} propose a symmetric Swin transformer-based architecture that maintains invertibility and topology preservation. All these ViT-based approaches mentioned previously improve registration performance by benefiting from long-range modeling information via the transformers. However, all these methods utilize the transformer to model coarse-grained information that may restrict the contributions of transformers in these models. Unlike these approaches, we present an unsupervised Swin transformer-based method for deformable image registration, which enhances the contributions of the Swin transformer blocks by recovering the feature representations and automatically builds the connections between windows.

\section{Methods}

\subsection{Image Registration}

\begin{figure*}[ht!]
    \centering
    \includegraphics[width=1.0\linewidth]{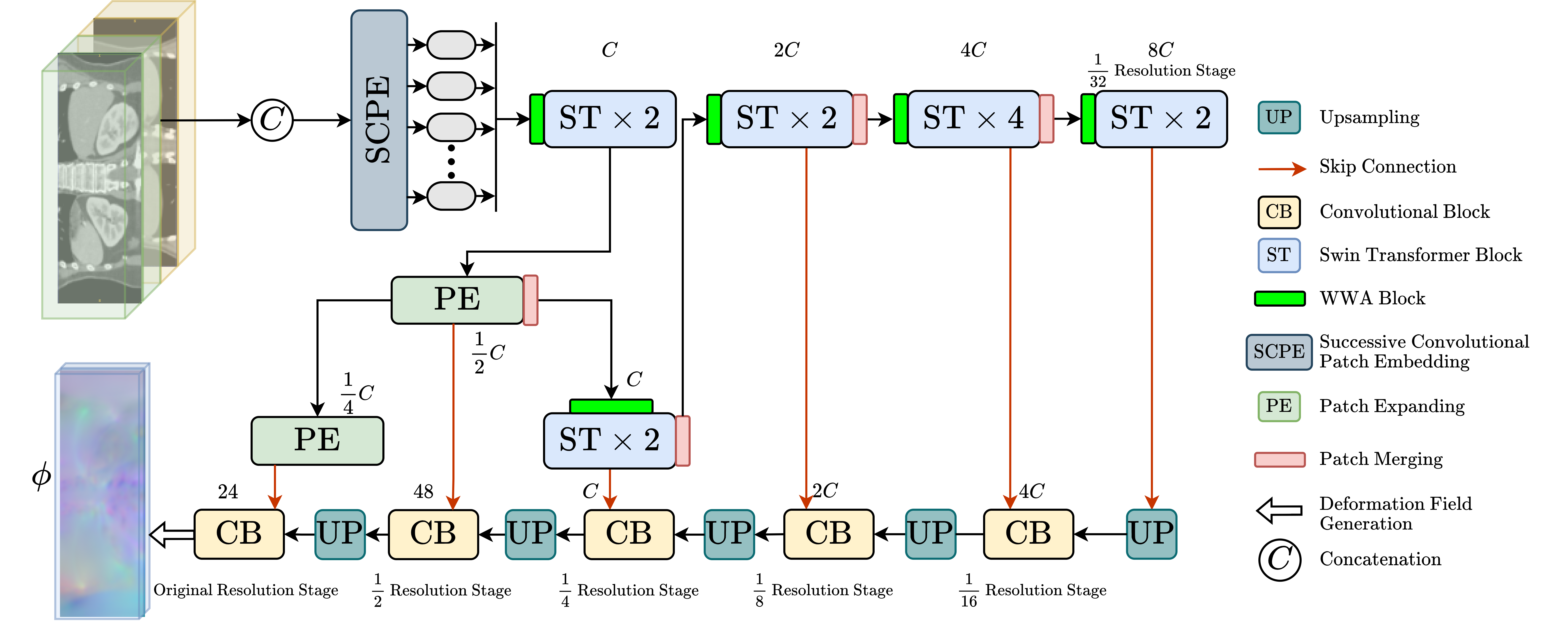}
    \caption{Overview of the RFR-WWANet. The parameter on each block indicates the number of channels output by that block. The input yellow and green cuboid images represent the moving and fixed image, and $\phi$ is the deformation field output from RFR-WWANet.}
    \label{fig:architechture}
\end{figure*}

Deformable image registration minimizes an energy function to establish a dense spatial correspondence between an image pair. Optimization aims to find an optimal deformation that can be formulated as

\begin{align}
        \hat{\phi}=\mathop{\arg\min}\limits_{\phi}(\mathcal{L}_{sim}({I_m}{{\circ}{\phi}},I_f)+{\lambda}\mathcal{L}_{reg}(\phi)),
            \label{eq:goal}
\end{align}

where the $I_m$ and $I_f$ denote the moving and fixed image, $I_m{\circ}{\phi}$ is the warped image transformed via the deformation field $\phi$. $\mathcal{{L}}_{sim}$ is the similarity matrix to estimate the similarity between $I_m{\circ}{\phi}$ and $I_f$. $\mathcal{L}_{reg}(\phi)$ is the regularization, which enforces the smoothness of the deformation field, and $\lambda$ is a hyperparameter used to balance similarity and smoothness. Hence, the optimal deformation field $\hat{\phi}$ is obtained.

In this work, we follow Eq. \ref{eq:goal} to perform deformable image registration. Mean squared error (MSE) is utilized as the similarity metric to evaluate the similarity between an image pair, i.e., $\mathcal{L}_{sim}= \text{MSE}({I_m}{\circ}{\phi}, I_f)$, where $I_m$ and $I_f$ are moving and fixed images, respectively. ${\circ}$ is the spatial transform network (STN) \cite{2015-STN}, and ${I_m}{\circ}{\phi}$ represents $I_m$ warped via a deformation field $\phi$. STN can warp an image with a deformation field in an interpolation manner. We utilize the diffusion regularizer \cite{2018-VM} on the spatial gradients of a deformation field $\phi$, where the gradients are computed by using differences between neighboring voxels. The regularizer is denoted as $\mathcal{L}_{reg}=\text{Diff}({\phi})$. Hence, the loss function in this work is $\mathcal{L}(I_m,I_f,{\phi})= \text{MSE}({I_m}{\circ}{\phi},{I_f})+\lambda{\text{Diff}}(\phi)$, where $\lambda$ is the hyperparameter that determines the trade-off between similarity and regularity. We optimize the parameters of RFR-WWANet by minimizing this loss function.

\subsection{Swin Transformer}

Here, we pithily introduce the Swin transformer. The Swin transformer is a hierarchical transformer that computes the self-attention within each window by utilizing regular and shifted window-based MSA mechanism \cite{Swintrans}. The partitioning operation in the Swin transformer splits an input feature according to the window size setting. Then it flattens the split feature on the batch dimension in units of windows. Given the input feature representations ${m^l}$ of layer $l$, consecutive Swin transformer blocks at the same resolution stage operate as follows:

\begin{align}
    &{{\hat{m}}^{l}} = \text{W-MSA}\left( {\text{LN}\left( {{{{m}}^{l - 1}}} \right)} \right) + {{m}}^{l},\nonumber\\
    &{{m}^{l+1}} = \text{MLP}\left( {\text{LN}\left( {{{\hat{{m}}}^{l}}} \right)} \right) + {{\hat{m}}^{l}},\nonumber\\
    &{{\hat{m}}^{l+1}} = \text{SW-MSA}\left( {\text{LN}\left( {{{{m}}^{l+1}}} \right)} \right) + {{m}}^{l+1}, \nonumber\\
    &{{m}^{l+2}} = \text{MLP}\left( {\text{LN}\left( {{{\hat{{m}}}^{l+1}}} \right)} \right) + {{\hat{{m}}}^{l+1}},
    \label{eq:swin}
\end{align}

where W-MSA and SW-MSA denote the window-based multihead self-attention under regular and shifted window partitioning, respectively, LN denotes the layer normalization, and MPL denotes the multi-layer perceptron module \cite{2017-attention}. The attention matrix within a window computed by the self mechanism is formulated as

\begin{equation}
    \text{Attention}(Q, K, V) = \text{SoftMax}(QK^T/\sqrt{d}+B)V,
    \label{eq.att}
\end{equation}

where ${Q, K, V}$ are query, key, and value matrices, and $B$ is the learnable relative positional encoding.



\subsection{Restoring Feature Resolution Encoder}


\begin{figure*}[ht]
    \centering
    \includegraphics[width=1.0\linewidth]{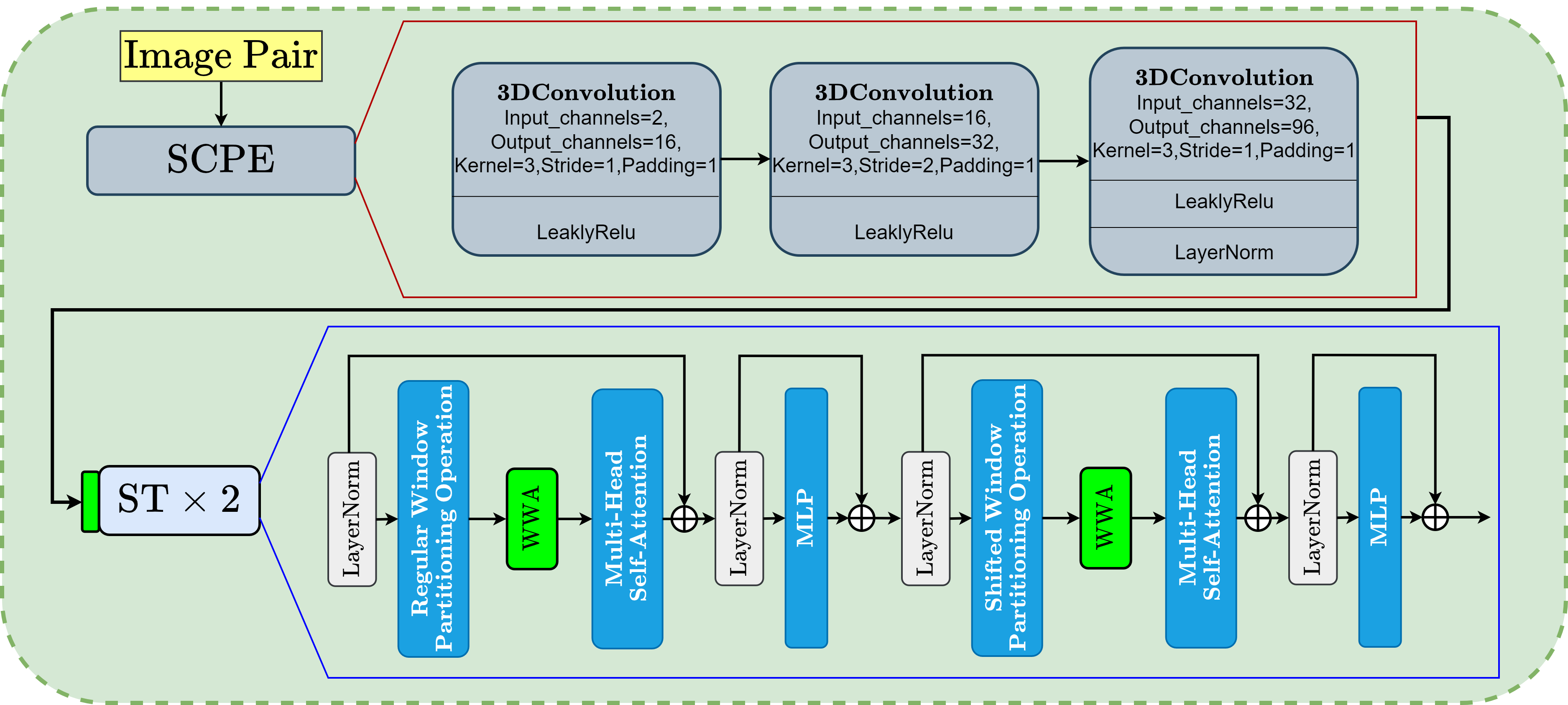}
    \caption{The details of SCPE and the details of two successive Swin transformer blocks with the proposed WWA blocks. The SCPE is utilized at the full resolution stage and extracts the feature representations to the 1/4 resolution stage. WWA blocks are exploited after the regular window partitioning operation and shifted window partitioning operation of two successive Swin transformer blocks.}
    \label{fig:supplement}
\end{figure*}

Let ${x}$ be a volume pair defined over a 3D spatial domain ${\mathcal{R}^{D{\times}H{\times}W{\times}2}}$ (i.e., ${x}\in{\mathcal{R}^{D{\times}H{\times}W{\times}2}}$), where $D$, $H$, and $W$ are the sizes of an image. As shown in Fig. \ref{fig:architechture}, the proposed RFR-WWANet is a U-shaped architecture. In this work, we utilize the successive convolutional patch embedding (SCPE) to obtain feature representations inputting into the Swin transformer block. SCPE consists of two convolutional blocks with a stride of $2$ and a kernel size of $3$ and one convolutional block with a stride of $1$ and a kernel size of $3$. SCPE outputs a sequence of 3D feature representations shape of $(\frac{D}{4},\frac{H}{4},\frac{W}{4},C)$, where $C$ is the number of channels.

After SCPE, We employ the previously introduced Swin transformer as the basic building block, which computes MSA locally in non-overlapping windows. This work uses the cuboid window shape of $(d,h,w)$ to adapt the input image shape that can be evenly divided by the image size. Thus, the number of partitioned windows $N$ is obtained by this formulation:
$N=\frac{D}{4d} \times \frac{H}{4h} \times \frac{W}{4w}$,
For the subsequent layer $l+1$, we adopt a 3D cyclic-shifting \cite{Swintrans} for efficient batch computation of the shifted windows. The partitioned windows are shifted by $(\lfloor \frac{d}{2}\rfloor, \lfloor \frac{h}{2}\rfloor, \lfloor \frac{w}{2}\rfloor)$ voxels. We employ the proposed WWA mechanism to bridge each window connection and relevance after a feature representation is transformed into a window sequence. SCPE and two successive WWA-based Swin transformer blocks are shown in Fig. \ref{fig:supplement}. Each Swin transformer block computes the attention matrix following Eq. \ref{eq:swin} and Eq. \ref{eq.att}.

The patch expanding operation is utilized in many U-shaped transformer-based approaches \cite{my,transmorph,2021-swinunet}. The patch expanding operations expand the feature maps along different channels, then reshape the feature representations into the shape of twice the input resolution and half the number of input channels, which achieves the recovery of the deep feature representations output from the Swin transformer block.
Specifically, first at the $1/4$ resolution stage, the Swin transformer block is utilized for modeling the feature representation output from SCPE. The patch expanding block recovers the resolution of this feature representation to the $1/2$ resolution stage. Next, this feature map is processed in upward and downward branches. The recovered feature representations are sent to the next Swin transformer block in the downward branch. The patch merging operation is utilized to reduce the feature size to the subsequent resolution stage by concatenating the features of each group of $2\times2\times2$ neighboring patches, then applying a linear operation to reduce the number of channels.
Simultaneously, the restored feature representation continues to be restored to the original resolution size stage. Skip connections connect the feature representations in the encoding with corresponding convolutional blocks in decoders at $1/4$, $1/2$ and original resolution stages. Since the restored feature representations connect to the higher resolution stage decoder blocks, we believe that the contributions of Swin transformer blocks in the first layer are improved.
At the remaining resolution stages of the encoder, we continue to use the Swin transformer blocks to model deep feature representations until the bottom is reached.


\subsection{Weighted Window Attention Mechanism} \label{WWA Example}

\begin{figure}[ht!]
    \centering
    \includegraphics[width=1.\linewidth]{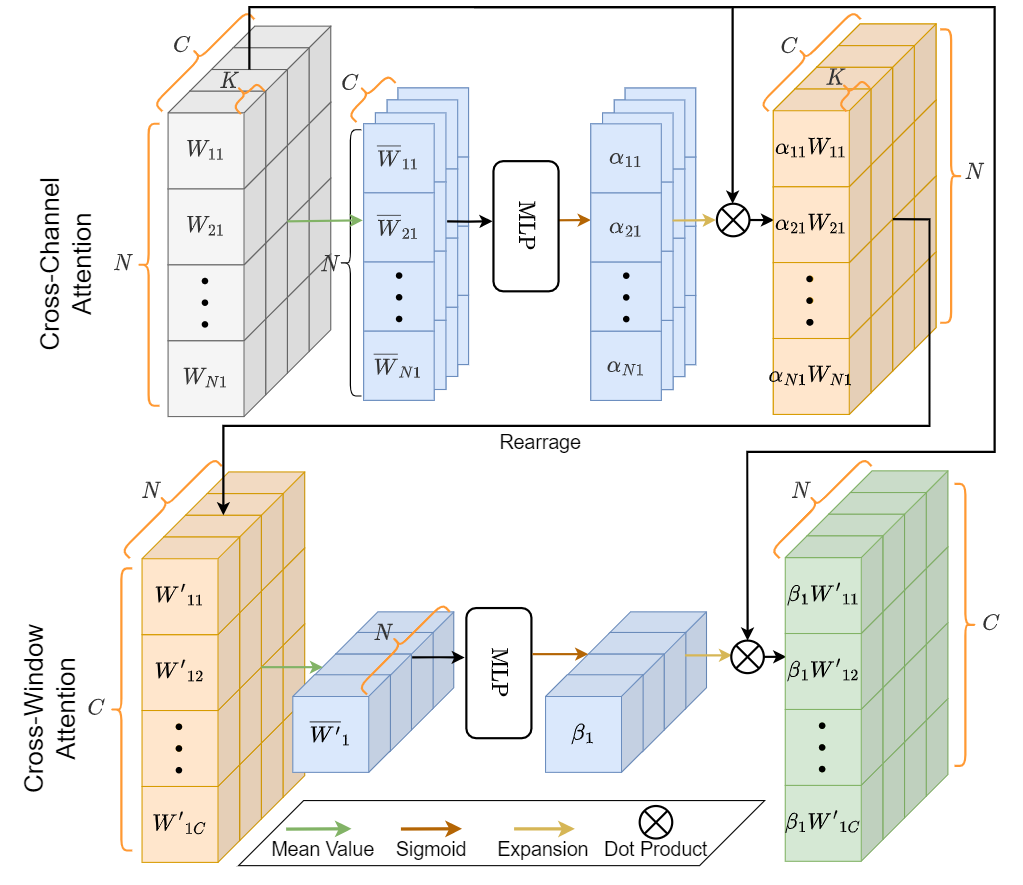}
    \caption{Overview of the proposed WWA block. $N$, $K$, and $C$ represent the number of windows, the number of elements in each window, and the number of channels in each window, respectively.}
    \label{fig:myattention}
\end{figure}
For the standard Swin transformer, a deep feature representation from an image pair is transformed into a sequence of windows on the batch direction by the regular or shifted window partitioning operation. Thus, the transformer can compute the attention matrix within a window. Inspired by \cite{ECA, CBAM}, to improve the capability of building interactions between windows, we propose WWA, an attention mechanism to compute attention weights between windows and allocate the weights for each window. The overview of WWA is shown in Fig. \ref{fig:myattention}.

Given a window sequence $W$ of size $(N\times K \times C)$, which is output from a regular and shifted window partitioning operation, where $N$ is the number of windows, $K$ is the length of each window, and $C$ is the number of channels. The window length of $K$ denotes the number of elements in $W_{ij}$. An element of $W$ can be denoted as $W_{ijk}$, where $i \in N, j\in C, k\in K$. As shown in Fig. \ref{fig:myattention}, $W_{ij}$ with $K$ elements is drawn as a sub-cube for a better view. In the cross-channel attention phase, an input window sequence is transformed into a matrix $\overline{W}$ of mean values computed by an average function. It can be formulated as follows:

\begin{align}
   \begin{matrix}
        \overline{W}_{ij}= (\sum_{k=1}^K W_{i,j,k}) / K, (i \in N, j \in C).
    \end{matrix}
\end{align}

The mean value of the elements within a window $W_{ij}$ is denoted as $\overline{W}_{ij}$. Then, the Sigmoid function is utilized after the MLP module. This MLP module has one hidden layer with a reduction factor of 4 to compute the cross-channel attention maps $\alpha$ of $\overline{W}$. Each element ${\alpha}_{ij}$ in the attention maps is expanded to the size of $(N\times K \times C)$. Finally, the weighted window sequence $W'$ is obtained by taking dot product $\alpha$ and $W$. In short, the cross-channel attention mechanism that computes $W'$ can be expressed as

\begin{align}
    W'=\text{Sigmoid}(\text{MLP}(\overline{W})) \otimes W.
\end{align}

\begin{figure}[t!]
    \centering
    \includegraphics[width=1.\linewidth]{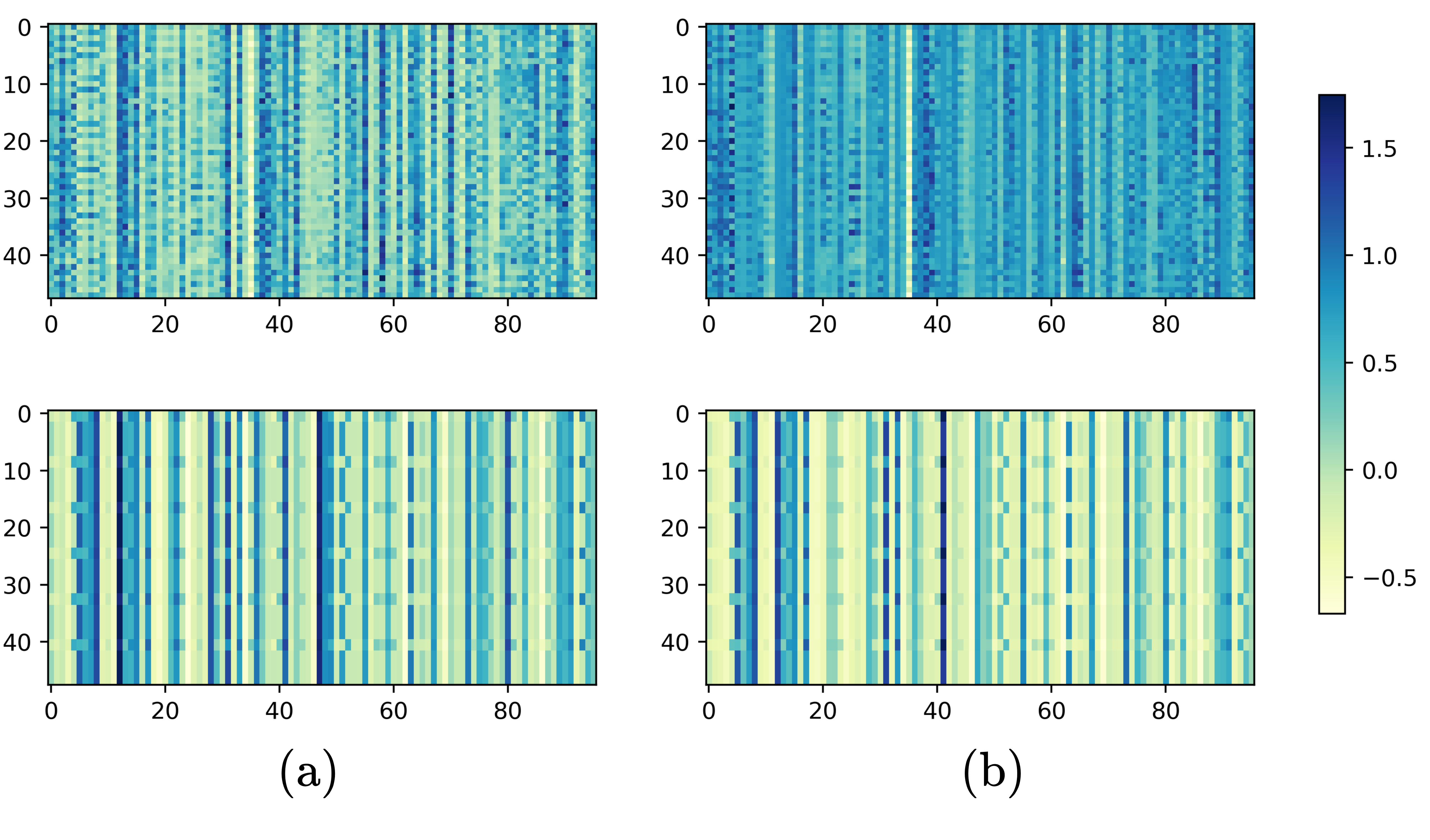}
        \caption{Two example slices of the window sequence input into and output from WWA block. Each image denotes the window with 96 channels (horizontal axis) and 48 elements (vertical axis) at the $1/4$ resolution stage. Column (a) represents a window before inputting into the WWA block, and column (b) represents the output from the WWA block.}
    \label{fig:attn_maps}
\end{figure}

In the cross-window attention phase, the input $W'$ from the previous cross-channel attention block is reshaped to the shape of $(K \times C \times N)$. Thus, a window of the window sequence is defined as $W'_{ij}$ and each window with $K$ elements inside. As shown in Fig. \ref{fig:myattention}, similar to the previously described, cross-window attention first computes the mean value $W'_{ij}$ along the channel direction to obtain $\overline{W'}_i$. This can be formulated as

\begin{align}
   \begin{matrix}
    \overline{W'}_i= (\sum_{j=1}^C W_{ij}) / C, (i \in N, j \in C).
    \end{matrix}
\end{align}

After the sequence of $\overline{W}'$ is obtained, an MLP with the same configuration mentioned above is utilized to compute the attention sequence $\beta$. Then, $\beta$ is expanded to the shape of $W'$, and dot product with $W$ to obtain the weighted windows $W''$, which can be expressed as follows:

\begin{align}
    W''=\text{Sigmoid}(\text{MLP}(\overline{W}_i))\otimes W.
\end{align}

Hence, the weighted windows $W''$ are computed, where these weights are computed from the information between windows. That builds the connections between windows. An example of a window input into and output from WWA is shown in Fig. \ref{fig:attn_maps}.
Given an upper abdominal image with the size of $198 \times 128 \times 64$, at the 1/4 resolution stage of RFR-WWANet, it is transformed into a feature representation size of $48 \times 32 \times 16 \times 96$, where 96 is the number of channels. Based on the window size configuration described in Section \ref{implemention}, a window contains $6 \times 4 \times 2 = 48$ elements. In the first Swin transformer block, the window partitioning operation is used to transform the feature representation into a window sequence with a shape of $(N, 48, 96)$, where $N = \frac{48}{6} \times \frac{32}{4} \times \frac{16}{2} = 512$. Two slices are randomly selected from the window sequence with 512 batches, resulting in an example of a window with 48 elements and 96 channels in Fig. \ref{fig:attn_maps}.

\section{Experiments}

\subsection{Dataset and Preprocessing}
We validate the proposed method for the atlas-based deformable abdominal CT scan registration task. Two publicly available datasets, WORD \cite{word} and BTCV \cite{BTCV}, are utilized for our experiments.

\textbf{WORD:} This dataset consists of 150 abdominal CT scans from 150 patients with 30495 slices. There are 16 organs with fine pixel-level annotations. Each CT volume in WORD consists of 159 to 330 slices of $512\times 512$ pixels. The in-plane resolution of each slice in BTCV is $0.976 \times 0.976 \ \text{mm}^2$, and the spacing of these slices ranges from 2.5 $\text{mm}$ to 3.0 $\text{mm}$. WORD contains three subsets, including 100 scans for training, 20 Scans for validation, and 30 scans for testing.

\textbf{BTCV:} This dataset consists of 50 abdominal CT scans, each scan with 13 organ annotations. Each volume contains 85 to 198 slices of $512 \times 512$ pixels. The in-plane resolution of each slice varies from $0.54 \times 0.54 \ \text{mm}^2$ to $0.98 \times 0.98 \ \text{mm}^2$, and the spacing of these slices ranges from 2.5 $\text{mm}$ to 5.0 $\text{mm}$. BTCV dataset is divided into two parts: one for training and the other for testing.

We select the training and validation dataset in WORD and the training dataset in BTCV. We choose these datasets in WORD and BTCV because these volumes have corresponding labels, which allows us to perform data preprocessing in the above manner. And through the labels corresponding to these data, we can compare the results of each baseline method with the results of RFR-WWANet.
We augment the number of volumes in the training dataset of WORD to 200 by utilizing random elastic transformation in TorchIO \cite{torchio}. This augmentation applies slight deformation to the volumes, preserving their original topology.
During preprocessing, we resample all volumes to a voxel spacing of $1.5 \times 1.5 \times 1.0 \ \text{mm}^3$. The intensity values are first clipped in the range of $[-200,300]$ Hounsfield Units and then normalized to the range of $[0,1]$. After that, we flip the volumes in BTCV to make it consistent with the coordinate direction of WORD.
We use the anatomically affine transformation in ANTs \cite{Ants} to preprocess the scans in BTCV and globally align them with the atlas in WORD.
We keep the segmentation maps of the liver, spleen, left kidney, right kidney, stomach, gallbladder, and pancreas in the labels of WROD and BTCV, then remove the rest. We use the retained segmentation maps of each label to find the largest and smallest locations of these organs in three dimensions, thereby cropping the upper abdominal image of interest. Each volume is then resampled into $192\times 128 \times 64$.

\subsection{Baseline Methods}

We compare the proposed RFR-WWANet with six deformable registration approaches, including four deep learning-based and two traditional methods. Two conventional methods are deedsBCV \cite{deeds} and SyN \cite{SyN}. These traditional methods use the recommended parameter settings. Four deep learning-based models include VoxelMorph \cite{2018-VM}, Vit-V-Net \cite{2021-vitV}, SymTrans \cite{my}, and TransMorph \cite{transmorph}. VoxelMorph is the pure CNN-based U-shaped model. Vit-V-Net first introduces ViT to the medical image registration task, which applies the ViT backbone at the bottom of the U-shaped architecture. SymTrans and TransMorph are the other ViT-based models. SymTrans utilizes convolution-based efficient MSA and builds the symmetric ViT-based registration model. TransMorph is the current state-of-the-art approach, which consists of a Swin transformer-based encoder and a convolution-based decoder. These four deep learning-based baseline methods and the proposed RFR-WWANet use the same loss functions, and the hyperparameter $\lambda=0.04$ is utilized for training these methods on the training set of WORD. We found that when the hyperparameter $\lambda=0.04$ for VoxelMorph, $\lambda=0.03$ for ViT-V-Net, $\lambda=0.02$ for TransMorph, $\lambda=0.03$ for SymTrans, the baseline methods perform best on the Dice metric. Experiments on hyperparameter $\lambda$ settings can be found in Section \ref{hyperparameters}.

\begin{figure*}[ht]
    \centering
\includegraphics[width=1.0\textwidth,scale=1.0]{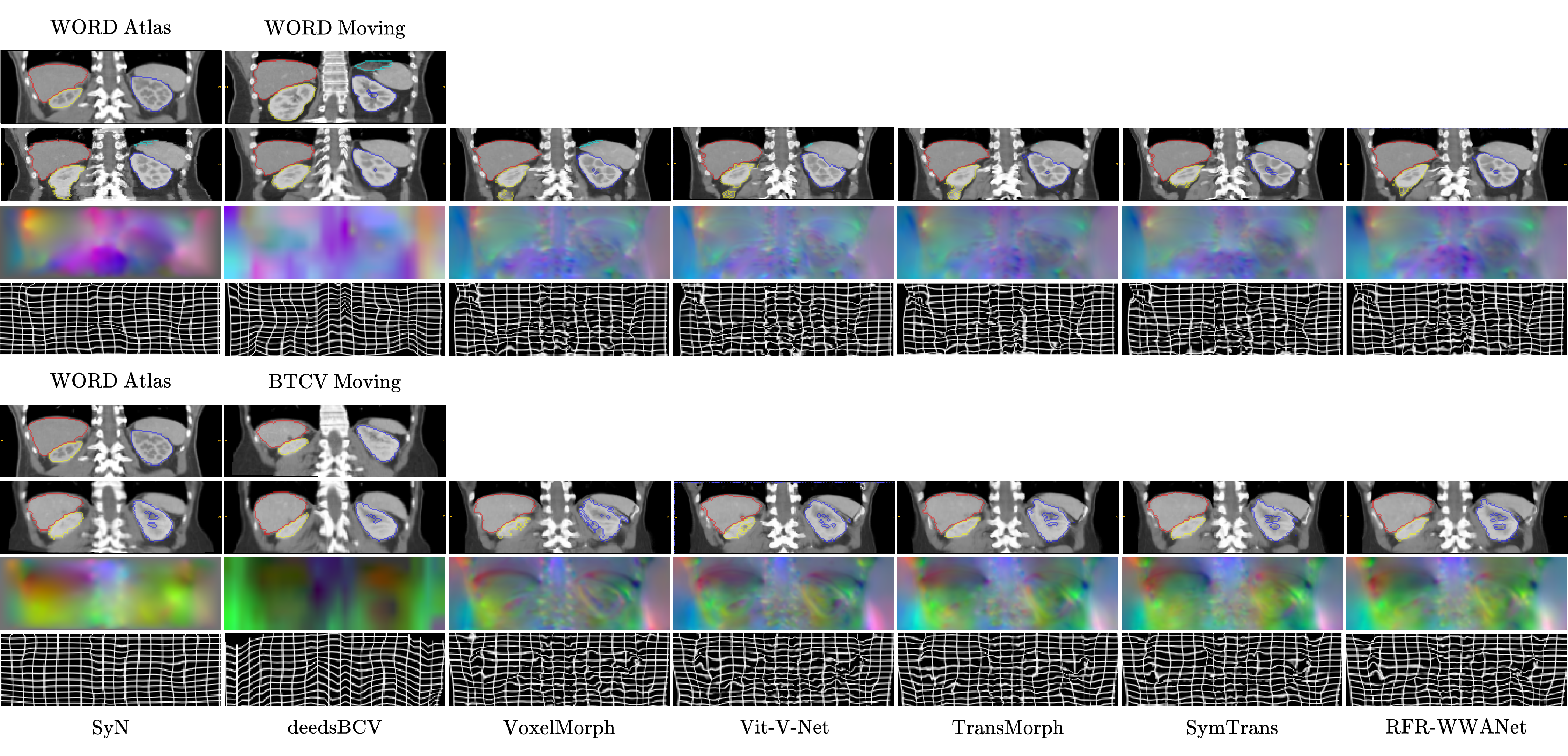}
    \caption{Images showing an example of a registration image pair. Four organs in each slice are the stomach, left and right kidneys, and liver. They are marked in light blue, dark blue, yellow, and red, respectively. In these two parts, the first row shows the input image pair, and the second row shows the warped image by different methods. Warped grids are used to observe deformations roughly. Each channel in the RGB image corresponds to a direction in the deformation field, where each pixel represents the displacement of the voxel at that position in three directions..}
    \label{fig:show}
\end{figure*}

\begin{table*}[htp!]
  \centering
       \caption{Comparison results of Dice (higher is better), $\text{HD}_{95}$ (lower is better), and \% $|J_{\phi}| \leq 0$ (lower is better) on the WORD validation dataset. These seven omitted words are Liv: liver, Spl: spleen, left and right kidneys: (Lkid, Rkid), stomach: Sto, gallbladder: Gall, and pancreas: Pan.}
  \resizebox{1.\textwidth}{!}{
    \begin{tabular}{crlllllll}
    \toprule
    \multicolumn{2}{c}{Method} & SyN   & deedsBCV & VoxelMorph & Vit-V-Net & TransMorph & SymTrans & RFR-WWANet \\
    \midrule
    \multicolumn{1}{c}{\multirow{8}[4]{*}{Dice}}
	& \multicolumn{1}{l}{Liv}	&0.863 $\pm$ 0.033	&0.860 $\pm$ 0.047	&0.881 $\pm$ 0.022	&0.888 $\pm$ 0.020 	        &0.894 $\pm$ 0.019 	&0.897 $\pm$ 0.018 	&\textbf{0.898 $\pm$ 0.018}\\
	& \multicolumn{1}{l}{Spl}	&0.732 $\pm$ 0.098	&0.693 $\pm$ 0.065	&0.734 $\pm$ 0.104	&0.747 $\pm$ 0.107 	        &0.765 $\pm$ 0.102 	&0.749 $\pm$ 0.107 	&\textbf{0.770 $\pm$ 0.096} \\
	& \multicolumn{1}{l}{Lkid}	&0.706 $\pm$ 0.159	&\textbf{0.774 $\pm$ 0.072}	&0.668 $\pm$ 0.156	&0.682 $\pm$ 0.157 	&0.685 $\pm$ 0.161 	&0.683 $\pm$ 0.171 	& 0.721 $\pm$ 0.165 \\
	& \multicolumn{1}{l}{Rkid}	&0.680 $\pm$ 0.119	&0.702 $\pm$ 0.089	&0.666 $\pm$ 0.104	&0.698 $\pm$ 0.094 	        &0.709 $\pm$ 0.112 	&0.714 $\pm$ 0.114 	&\textbf{0.734 $\pm$ 0.104} \\
	& \multicolumn{1}{l}{Sto}	&0.462 $\pm$ 0.128	&0.479 $\pm$ 0.113	&0.508 $\pm$ 0.113	&0.520 $\pm$ 0.106 	        &0.532 $\pm$ 0.106 	&0.528 $\pm$ 0.110 	&\textbf{0.535 $\pm$ 0.104} \\
	& \multicolumn{1}{l}{Gall}	&0.139 $\pm$ 0.151	&0.161 $\pm$ 0.163	&0.183 $\pm$ 0.173	&0.183 $\pm$ 0.165 	        &0.188 $\pm$ 0.187 	&0.199 $\pm$ 0.182 	&\textbf{0.217 $\pm$ 0.198} \\
	& \multicolumn{1}{l}{Pan}	&0.418 $\pm$ 0.118	&0.401 $\pm$ 0.118	&0.393 $\pm$ 0.129	&0.397 $\pm$ 0.132 	& \textbf{0.420 $\pm$ 0.134} 	&0.410 $\pm$ 0.139 	&0.406 $\pm$ 0.133 \\
\cmidrule{2-9}        & \multicolumn{1}{l}{Avg.} & 0.571 $\pm$ 0.066 & 0.582 $\pm$ 0.250 & 0.556 $\pm$ 0.072 & 0.588 $\pm$ 0.071 & 0.599 $\pm$ 0.078 & 0.597 $\pm$ 0.081 & \textbf{0.612 $\pm$ 0.077} \\
    \midrule
    \multicolumn{1}{c}{\multirow{8}[4]{*}{$\text{HD}_{95}$}}
    &\multicolumn{1}{l}{Liv}&11.869 $\pm$ 5.740&10.841 $\pm$ 6.391 &10.457 $\pm$ 4.999 &10.339 $\pm$ 4.773 &\textbf{10.286 $\pm$ 4.794} &10.458 $\pm$ 4.822 &10.509 $\pm$ 5.098\\
    &\multicolumn{1}{l}{Spl}&11.292 $\pm$ 5.939&13.011 $\pm$ 6.557 &11.191 $\pm$ 6.906&11.319 $\pm$ 7.249 &10.946 $\pm$ 6.518 &11.394 $\pm$ 7.226 &\textbf{10.465 $\pm$6.423}\\
    &\multicolumn{1}{l}{Lkid}&8.314 $\pm$ 4.315&\textbf{4.878 $\pm$ 1.498} &10.004 $\pm$ 4.246&10.243 $\pm$ 4.368 &10.748 $\pm$ 4.369 &10.639 $\pm$ 4.717 &9.428 $\pm$ 4.438\\
    &\multicolumn{1}{l}{Rkid}&10.665 $\pm$ 4.087&\textbf{8.486 $\pm$ 3.209} &10.743 $\pm$ 3.258&8.979 $\pm$ 2.666 &9.322 $\pm$ 2.830 &9.297 $\pm$ 3.530 &8.801 $\pm$ 3.335\\
    &\multicolumn{1}{l}{Sto}&19.462 $\pm$ 7.064&18.694 $\pm$ 7.564 &17.151 $\pm$ 6.970&16.793 $\pm$ 7.148 &16.978 $\pm$ 6.898 &16.572 $\pm$ 7.219 &\textbf{16.362 $\pm$ 6.861}\\
    &\multicolumn{1}{l}{Gall}&18.061 $\pm$ 7.523&18.179 $\pm$ 8.394 &16.042 $\pm$ 6.575&15.784 $\pm$ 6.465 &15.991 $\pm$ 6.831 &15.816 $\pm$ 6.254 &\textbf{15.534 $\pm$ 7.234}\\
    &\multicolumn{1}{l}{Pan}&\textbf{10.718 $\pm$ 5.278}&12.072 $\pm$ 4.639 &11.124 $\pm$ 4.109&11.370 $\pm$ 4.2661 &10.852 $\pm$ 3.904 &11.103 $\pm$ 4.199 &10.878 $\pm$ 4.029\\

\cmidrule{2-9}
          & \multicolumn{1}{l}{Avg.} & 12.912 $\pm$ 3.798 & 12.309 $\pm$ 7.496 & 12.877 $\pm$ 3.628 & 12.118 $\pm$ 3.696 & 12.160 $\pm$ 3.7031 & 12.183 $\pm$ 3.844 & \textbf{11.711 $\pm$ 3.826} \\
    \midrule

    \% $|J_{\phi}| \leq 0$   &       & $3.271e^{-3}$ $\pm$ $1.391e^{-2}$ & $2.640e^{-2} \pm 3.467e^{-2}$ & 3.080 $\pm$ 0.962 & 1.818 $\pm$ 0.548 & 2.583 $\pm$ 0.699 & 1.963 $\pm$ 0.718 & 1.166 $\pm$ 0.465 \\
    \bottomrule
    \end{tabular}%
    }
  \label{tab:two_dataset_results}%
\end{table*}%

\subsection{Evaluation Metrics}

We use the Dice score and Hausdorff Distance \cite{hausdorff} to evaluate the registration accuracy. The Dice score is a metric that calculates the overlap between the ground truth segmentation maps and the warped moving image corresponding segmentation maps. The Dice metric is the most widely used metric in unsupervised medical image registration research. The Hausdorff Distance calculates surface distances between the warped and ground-truth labels. $HD_{95}$ calculates the $95th$ percentile of surface distances between them. Nonpositive Jacobian determinant $|J_{\phi}| \leq 0$ is utilized to calculate the number of folding in a deformation field.

To test whether our proposed method improves significantly over the baseline methods, we perform the paired t-test on the pairs consisting of the experimental results of RFR-WWANet and the experimental result of each baseline method.

\subsection{Implementation Details} \label{implemention}
The proposed framework RFR-WWANet is implemented by using PyTorch \cite{pytorch}. We set the regularization parameter $\lambda$ to 0.04. We employ the Adam optimizer to optimize the parameters of the proposed network, with a learning rate of 1e-4, on an NVIDIA RTX3080 10 GB GPU. The maximum number of training epochs for the RFR-WWANet and baseline methods is 300. RFR-WWANet is implemented following Fig. \ref{fig:architechture}, where the number of channels $C$ is set to 96. The window size in this work is set to $(6,4,2)$. The number of heads of the WWA-based Swin transformer blocks is $(4,4,8,8)$.

\subsection{Experimental Results}

\begin{table*}[ht!]
  \centering
      \caption{Comparison results of Dice (higher is better), $\text{HD}_{95}$ (lower is better), and \% $|J_{\phi}| \leq 0$ (lower is better) on the BTCV validation dataset. These seven omitted words are Liv: liver, Spl: spleen, left and right kidneys: (Lkid, Rkid), stomach: Sto, gallbladder: Gall, and pancreas: Pan.}
    \resizebox{1.\textwidth}{!}{
    \begin{tabular}{lrlllllll}
    \toprule
    \multicolumn{2}{c}{Method} & SyN   & deedsBCV & VoxelMorph & Vit-V-Net & TransMorph & SymTrans & RFR-WWANet \\
    \midrule
    \multicolumn{1}{c}{\multirow{8}[4]{*}{Dice}}
    &\multicolumn{1}{l}{Liv}&0.818 $\pm$ 0.044&0.852 $\pm$ 0.035 &0.845 $\pm$ 0.036&0.850 $\pm$ 0.036 &0.861 $\pm$ 0.034 &\textbf{0.895 $\pm$ 0.032} &0.869 $\pm$ 0.031\\
    &\multicolumn{1}{l}{Spl}&0.659 $\pm$ 0.135&0.635 $\pm$ 0.151 &0.694 $\pm$ 0.101&0.697 $\pm$ 0.104 &0.716 $\pm$ 0.091 &0.699 $\pm$ 0.112 &\textbf{0.720 $\pm$ 0.101}\\
    &\multicolumn{1}{l}{Lkid}&0.579 $\pm$ 0.157&0.586 $\pm$ 0.141 &0.567 $\pm$ 0.152&0.570 $\pm$ 0.170 &0.581 $\pm$ 0.172 &0.574 $\pm$ 0.187 &\textbf{0.608 $\pm$ 0.161}\\
    &\multicolumn{1}{l}{Rkid}&0.596 $\pm$ 0.150&0.669 $\pm$ 0.092 &0.619 $\pm$ 0.143&0.643 $\pm$ 0.137 &\textbf{0.672 $\pm$ 0.147} &0.654 $\pm$ 0.166 &0.665 $\pm$ 0.157\\
    &\multicolumn{1}{l}{Sto}&0.349 $\pm$ 0.103&\textbf{0.468 $\pm$ 0.120} &0.391 $\pm$ 0.117&0.387 $\pm$ 0.124 &0.375 $\pm$ 0.123 &0.369 $\pm$ 0.126 &0.388 $\pm$ 0.124\\
    &\multicolumn{1}{l}{Gall}&0.209 $\pm$ 0.189&0.271 $\pm$ 0.182 &0.265 $\pm$ 0.177&0.278 $\pm$ 0.179 &0.283 $\pm$ 0.191 &\textbf{0.293 $\pm$ 0.161} &{0.281 $\pm$ 0.170}\\
    &\multicolumn{1}{l}{Pan}&0.289 $\pm$ 0.076&\textbf{0.328 $\pm$ 0.102} &0.291 $\pm$ 0.096&0.287 $\pm$ 0.097 &0.291 $\pm$ 0.109 &0.295 $\pm$ 0.104 &0.306 $\pm$ 0.105\\

\cmidrule{2-9}
& \multicolumn{1}{l}{Avg.}
& 0.500 $\pm$ 0.070 & 0.544 $\pm$ 0.226 & 0.518 $\pm$ 0.065 & 0.530 $\pm$ 0.068 & 0.540 $\pm$ 0.076 & 0.535 $\pm$ 0.078 & \textbf{0.548 $\pm$ 0.071} \\
    \midrule
    \multicolumn{1}{c}{\multirow{8}[4]{*}{$\text{HD}_{95}$}}
&\multicolumn{1}{l}{Liv}&12.062 $\pm$ 4.731&10.327 $\pm$ 3.189 &10.827 $\pm$ 4.046&10.524 $\pm$ 4.106 &10.537 $\pm$ 3.540 &11.021 $\pm$ 3.855 &\textbf{10.301 $\pm$ 3.587} \\
&\multicolumn{1}{l}{Spl}&11.246 $\pm$ 5.241&11.614 $\pm$ 5.885 &11.222 $\pm$ 3.777&11.321 $\pm$ 4.397 &11.849 $\pm$ 4.645 &11.543 $\pm$ 4.438 &\textbf{11.069 $\pm$4.535} \\
&\multicolumn{1}{l}{Lkid}&9.910 $\pm$ 3.685&\textbf{9.319 $\pm$ 4.301} &12.200 $\pm$ 4.955&12.768 $\pm$ 5.425 &12.831 $\pm$ 5.439 &12.502 $\pm$ 5.123 &12.116 $\pm$ 5.041\\
&\multicolumn{1}{l}{Rkid}&10.193 $\pm$ 4.659&\textbf{7.710 $\pm$ 3.189} &9.413 $\pm$ 3.180&8.928 $\pm$ 3.163 &8.415 $\pm$ 3.130 &8.925 $\pm$ 3.851 &8.653 $\pm$ 3.832\\
&\multicolumn{1}{l}{Sto}&16.653 $\pm$ 3.937&15.809 $\pm$ 4.673 &14.567 $\pm$ 4.017&14.558 $\pm$ 4.420 &15.082 $\pm$ 4.270 &14.473 $\pm$ 4.561 &\textbf{14.003 $\pm$ 4.023}\\
&\multicolumn{1}{l}{Gall}&15.262 $\pm$ 6.029&15.441 $\pm$ 6.989 &14.624 $\pm$ 4.978&14.787 $\pm$ 4.843 &14.473 $\pm$ 4.494 &14.191 $\pm$ 4.785 &\textbf{14.042 $\pm$ 4.089}\\
&\multicolumn{1}{l}{Pan}&\textbf{13.097 $\pm$ 3.330}&14.721 $\pm$ 4.251 &13.999 $\pm$ 3.417 & 13.909 $\pm$ 3.670 & 14.239 $\pm$ 3.304 &14.298 $\pm$ 3.542 &13.793 $\pm$ 3.654\\
\cmidrule{2-9}          & \multicolumn{1}{l}{Avg.} & 12.632 $\pm$ 2.117 & 12.061 $\pm$ 2.425 & 12.651 $\pm$ 2.451 & 12.407 $\pm$ 2.158 & 12.296 $\pm$ 2.316 & 12.252 $\pm$ 2.449 & \textbf{11.997 $\pm$ 2.300} \\
    \midrule

    \% $|J_{\phi}| \leq 0$   &       & $1.687e^{-3} \pm 7.34e^{-3}$ & $4.27e^{-2} \pm 1.59e^{-2}$ & 3.875 $\pm$ 0.861 & 2.478 $\pm$ 0.653 & 0.320 $\pm$ 0.716  & 0.264 $\pm$ 0.715  & 1.584 $\pm$ 0.461 \\
    \bottomrule
    \end{tabular}%
    }
  \label{tab:BTCV_result}%
\end{table*}%

We demonstrate the experimental results in two parts: the results using the WORD validation dataset and the results using the BTCV training dataset for testing. These two datasets are utilized to perform the atlas-based registration. The atlas is selected in the validation set of WORD, and the CT scan numbered 0001, is treated as the atlas image. Seven organs in the upper abdominal registration results are evaluated.

\textbf{Testing on WORD dataset.}
Table \ref{tab:two_dataset_results} shows the quantitative results of atlas-based registration. The calculated average result of each organ and the average results of these organs demonstrate that the proposed method, RFR-WWANet, achieves the highest Dice scores and the lowest Hausdorff Distance than the baseline methods. By comparison, we can find that all transformer-based methods outperform the CNN-based method, VoxelMorph. On the Dice metric, RFR-WWANet outperforms the second TransMorph by 1.3\% and the third SymTrans by 1.5\% on the average Dice score of 7 organs. On average, RFR-WWANet also achieves the best results on the Hausdorff Distance metric, which indicates that the segmentation maps transformed by RFR-WWANet can better match the segmentation maps of the atlas scan. The results of Vit-V-Net on abdominal images with significant structural differences are not as good as other transformer-based methods because it only applies the ViT blocks at the bottom of the model, so the ViT blocks can only model coarse-grained semantic information, resulting in poor registration performance. \%$|J_{\phi}| \leq 0$ denotes the proportion of the number of folding in a deformation field. Except for the two conventional methods that generate almost zero folding, the percentage folding of the remaining methods is basically at the same level.

Fig. \ref{fig:show} shows the qualitative results of the sample slices. We select slices containing the liver, spleen, left and right kidneys for visualization. We find that deedsBCV and three transformer-based methods, Vit-V-Net, TransMorph, and SymTrans, warp image well because, in this view, they are able to move the stomach out of the current slice. In this slice, both the transformed labels of the right kidney from VoxelMorph and Vit-V-Net lose their topological properties.

\textbf{Testing on BTCV dataset.}
Generally, the abdominal cavity dataset contains a small number of CT images. Although WORD contains 100 images for training and 30 for validation, more is needed to demonstrate the performance of the proposed method. Therefore, we use the BTCV dataset as the test set to test the performance of the baseline and the proposed methods. The Dice metric of all methods decreases, and the value of Hausdorff Distance also increases, although the BTCV dataset has undergone a preliminary affine transformation before testing. Except for Vit-V-Net, the other Dice score of the deep learning-based methods drop by about 6\%. RFR-WWANet still achieves the best results on the average Dice and the average $HD_{95}$ metrics. deedsBCV yields competitive average Dice scores for RFR-WWANet because they are non-learning algorithms that iteratively optimize the similarity between image pairs each time the deformation field is computed. So whether the dataset has been seen to these methods has no effect.

The second part of Fig. \ref{fig:show} shows the qualitative comparison of testing on the BTCV dataset. For these data that have never been seen, transformer-based methods still perform well than  VoxelMorph. By looking at the visualization of these sections, we note that the deformation of the VoxelMorph leads to partial destruction of the topological properties, i.e., the structures of the left and right kidneys are severely destroyed in the images.

\subsection{Hyperparameter Setting for Deep-learning approaches}
\label{hyperparameters}
\begin{figure}
    \centering
    \includegraphics[width=1.0\linewidth]{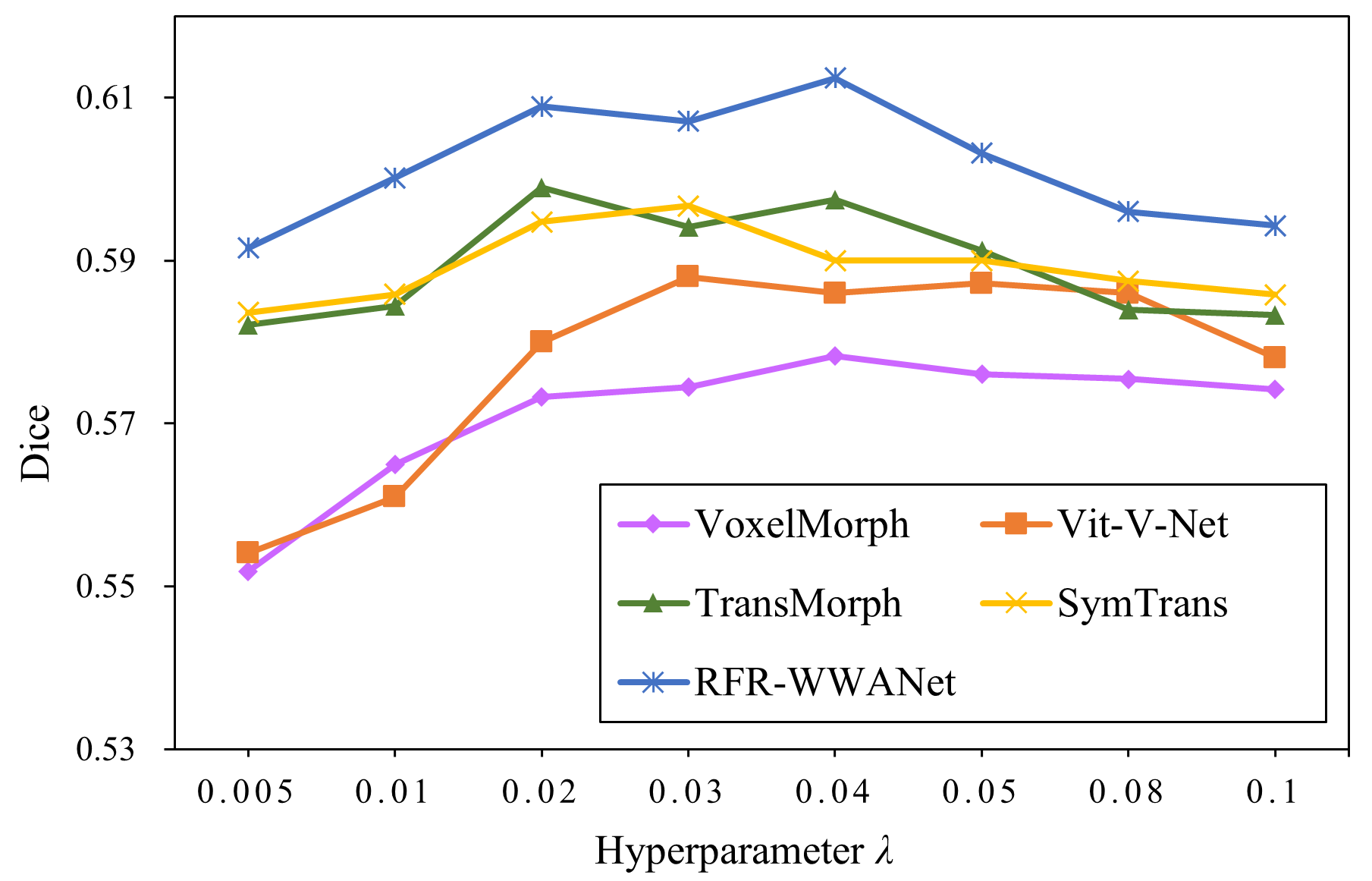}
    \caption{Dice score of WORD dataset for deep learning-based methods with varied hyperparameter settings.}
    \label{fig:my_label}
\end{figure}

In the deep learning-based registration model, the global regularization term and the similarity loss function jointly guide the learning of model weights. Therefore, the setting of the hyperparameter $\lambda$ of the global regularization term has a certain influence on the accuracy of the model. We use a grid search strategy to find $\lambda$ at which deep learning-based baseline methods and RFR-WWANet achieve the highest accuracy. The hyperparameters of all methods are specifically set to [0.005, 0.01, 0.02, 0.03, 0.04, 0.05, 0.08, 0.1]. We display the model accuracy using the corresponding $\lambda$ in the form of a line chart. Fig \ref{hyperparameters} shows that the accuracy of the baseline methods and RFR-WWANet is the highest when the hyperparameter $\lambda$ is set to 0.04 for VoxelMorph, 0.03 for Vit-V-Net, 0.02 for TransMorph, 0.03 for Symtrans, and 0.04 for our method.

\subsection{Computational Complexity}

\begin{table}[ht!]\small
  \centering
      \caption{Comparison of computational complexity between transformer-based methods. FLOP: the number of floating point operations.}
      \setlength{\tabcolsep}{5mm}{
    \begin{tabular}{lrr}
    \toprule
    Method & \multicolumn{1}{l}{Parameters (M)} & \multicolumn{1}{l}{FLOPs (G)} \\
    \midrule
    Vit-V-Net & 31.507 & 175.357 \\
    TransMorph & 46.689  & 300.645  \\
    SymTrans & 16.050 & 120.056 \\
    RFR-WWANet & 47.990 & 397.547 \\
    \bottomrule
    \end{tabular}}
      \label{tab:complexity}%
\end{table}%

Since the parameters of CNN-based models are usually much less than transformer-based models, we report the parameters of transformer-based models here. The parameters of four transformer-based models are shown in Table \ref{tab:complexity}. SymTrans is an approach optimized for model computational cost, so it contains the least parameters and FlOPs. Since ViT models are utilized at the bottom (i.e., 1/32 resolution level) of Vit-V-Net, resulting it has fewer parameters and FLOPs than TransMorph and RFR-WWANet.
Compared with TransMorph, which is also based on the Swin transformer, our model has 1.4 M more parameters and 96.9 G more FLOPs. Since two MLP layers are exploited in each WWA block, this leads to a considerable increase in the number of parameters and FLOPs. Although our method has a larger model size and FLOPs, our method only increases the parameters of TransMorph by 2.7\% and yields a more significant improvement in registration accuracy. This suggests that our method can achieve a more competitive registration quality with fewer additional computational resources.

\subsection{Additional Experiments on Brain MRI Dataset}

To measure the accuracy of our method on other organs, we additionally evaluate the deep learning-based baseline methods and the proposed RFR-WWANet on the brain MRI dataset, OASIS\cite{OASIS}, which is a dataset widely used in the deep learning-based registration research. There are 35 anatomical segmentation maps of each image in OASIS. The brain dataset is preprocessed using FreeSurfer\cite{FreeSurfer} according to the standard preprocessing process. Preprocessing includes affine transformation, skull stripping, resampling, etc. The shape of the brain MRI images after preprocessing is $96\times112\times96$. We perform the atlas-based registration task on the brain dataset. Four images are randomly selected as atlases, then 200 images are randomly selected as the training set, 30 as the validation set, and 50 as the testing set. The hyperparameters of all deep learning-based methods are set to 0.02, which is consistent with the setting of hyperparameters in the baseline methods when utilizing MSE as the similarity loss function.

\begin{table}[ht!]
\small
  \centering
  \caption{Comparison results of the brain dataset on Dice metric (higher is better) and \% $|J_{\phi}| \leq 0$ (lower is better).}

    \begin{tabular}{lll}
    \toprule
    Method & Dice  & \% $|J_{\phi}| \leq 0$ \\
    \midrule
    Affine Only & 0.591 $\pm$ 0.048 & -- \\
    SyN & 0.715 $\pm$ 0.500 & $4.747e^{-6} \pm 1.130e^{-4}$ \\
    deedsBCV & 0.693 $\pm$ 0.018 & $3.308e^{-6} \pm 4.182e^{-4}$ \\
    VoxelMorph & 0.729 $\pm$ 0.018 & $0.156 \pm 0.053$ \\
    Vit-V-Net &    0.730 $\pm$ 0.014  &  $ 0.136 \pm 0.043 $ \\
    TransMorph & 0.738 $\pm$ 0.017 & $ 0.164 \pm 0.048 $ \\
    SymTrans & \textbf{0.745 $\pm$ 0.017} & $ 0.169 \pm 0.046$ \\
    RFR-WWANet & 0.742 $\pm$ 0.016 & $ 0.167 \pm 0.047 $ \\
    \bottomrule
    \end{tabular}%
  \label{tab:brain}%
\end{table}%

Table \ref{tab:brain} shows the testing results on the baseline methods and RFR-WWANet. The proposed RFR-WWANet is on the second rank, less 0.3\% than the first rank method, SymTrans,  on the Dice metric. Compared with TransMorph on the third rank,  which is the Swin transformer-based approach, our method outperforms it by 0.4\% on the Dice metric. Compared with the CT image dataset of the upper abdominal dataset, the preprocessing of the MRI image of the brain can relatively strictly align the anatomical structures of each image. This can be seen from the ``Affine Only'' result after preprocessing. We report the ``Affine Only'' results in the abdominal experiment here: the Dice metric between the atlas scan and validation set is 0.408, and the standard deviation is 0.229. Especially the standard deviation of 0.229 is much greater than the value of 0.048 of the brain dataset, indicating that the anatomical structures could not be aligned well in abdominal CT scans. Therefore, SymTrans, a convolution-based self-attention method, can outperform TransMorph and RFR-WWANet on the brain dataset but not on the abdominal dataset. This suggests that modeling an abdominal CT image pair over the larger receptive distance improves the performance of the registration model.

\subsection{Significance Analysis}

To assess the significance between the baseline methods and RFR-WWANet, we conduct paired t-test and report the p-values on the abdominal CT registration task and the brain MRI registration task in Table \ref{tab:p-value}. Table \ref{tab:p-value} shows that the p-values are less than 0.05, even less than 0.01, which means that the improvement of RFR-WWANet is statistically significant.

\begin{table}[htbp]
\small
  \centering
  \caption{Calculated p-values between the results of each baseline method and RFR-WWANet on the abdominal CT and brain MRI datasets by paired t-test, respectively.}
   \setlength{\tabcolsep}{5mm}{
   \begin{tabular}{llr}
    \toprule
       Method   & Abdominal CT & \multicolumn{1}{l}{Brain MRI} \\
    \midrule
    SyN   & $5.754e^{-8}$ & $1.589e^{-33}$ \\
    deedsBCV & $1.787e^{-3}$ &  $3.405e^{-38}$ \\
    VoxelMorph & $3.862e^{-21}$ & $1.069e^{-18}$ \\
    Vit-V-Net & $8.383e^{-12}$      & $2.340e^{-22}$ \\
    TransMorph & $5.083e^{-12}$ & $1.229e^{-8}$ \\
    SymTrans & $3.521e^{-11}$ & $3.912e^{-3}$ \\
    \bottomrule
    \end{tabular}}%
  \label{tab:p-value}%
\end{table}%

\section{Ablation Studies}

\begin{figure*}[tp!]
    \centering
    \includegraphics[width=.95\linewidth]{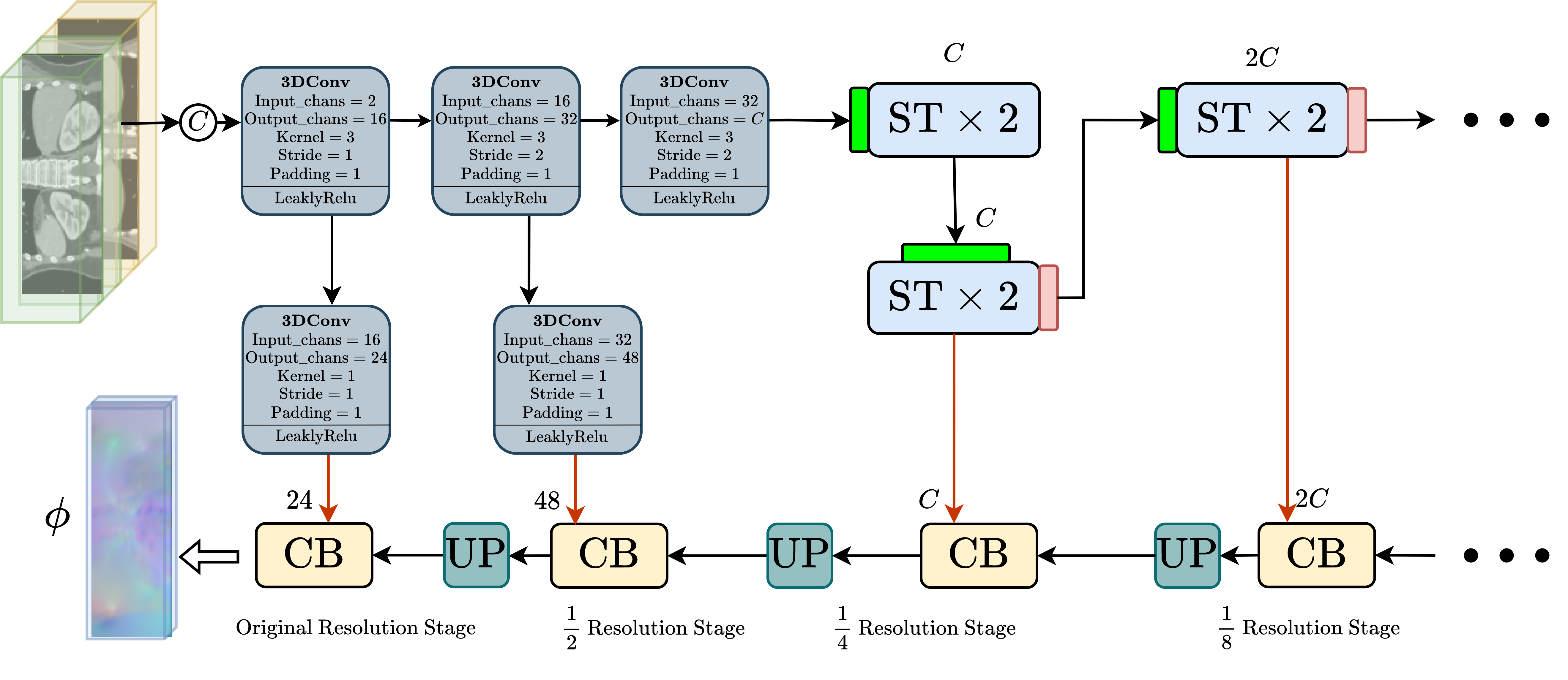}
    \caption{The ablation network of removing the recovery branch (i.e., R.B.) in RFR-WWANet.}
    \label{fig:arch_ablation}
\end{figure*}

We first remove the recovery branch and WWA to investigate the performance of the basic model, which is denoted as ``w/o R.B. and WWA''. We then remove the recovery branch and WWA block, denoted as ``w/o R.B.'' and ``w/o WWA'' in RFR-WWANet, respectively. For removing the recovery branch, two patch expanding layers shown in Fig. \ref{fig:architechture} are replaced with two convolutional layers with a kernel size of 1 and a stride of 1. These convolutional layers are used to compress the number of channels of feature representations output from SCPE so that the number of channels of the output feature representations can match the convolutional blocks in the decoder.
Note that ``w/o R.B.'' and ``w/o R.B. and WWA'' are conducted using additional convolutional layers with a kernel size of 1. Therefore, the parameters and FLOPs of these additional convolutional layers are not counted.

\begin{table}[ht!]
\small
  \centering
      \caption{Results of removing different components in RFR-WWANet. R.B. denotes the recovery branch. FLOP: the number of floating point operations.}
    \begin{tabular}{lrrr}
    \toprule
    Model & \multicolumn{1}{l}{Dice} & \multicolumn{1}{l}{Parameters (M)} & \multicolumn{1}{l}{FLOPs(G)} \\
    \midrule
    w/o R.B. and WWA &  0.600     & 47.060  & 395.734  \\
    w/o R.B          &  0.602     & 47.945  & 395.736  \\

    w/o WWA          &  0.607     & 47.106  & 397.546  \\
    RFR-WWANet         &  0.612     & 47.990  & 397.547  \\
    \bottomrule
    \end{tabular}%
  \label{tab:ablation}%
\end{table}%

The comparison results shown in Table \ref{tab:ablation} demonstrate that both the recovery branch and WWA are effective. Comparing ``w/o R.B. and WWA'' with ``w/o WWA'', these results indicate that the recovery branch generates few parameters and FLOPs. The comparison between `` w/o R.B. and WWA'' and ``w/o R.B '' indicates that the WWA blocks increase the number of parameters and FLOPs by almost 0.9 M and 0.001 G, respectively. Combined with the results on the Dice metric and these comparisons, it is proved that the recovery branch and WWA can effectively improve the performance of the registration model while generating a small number of parameters and FLOPs.

\begin{figure*}[ht!]
    \centering
    \includegraphics[width=1.\linewidth]{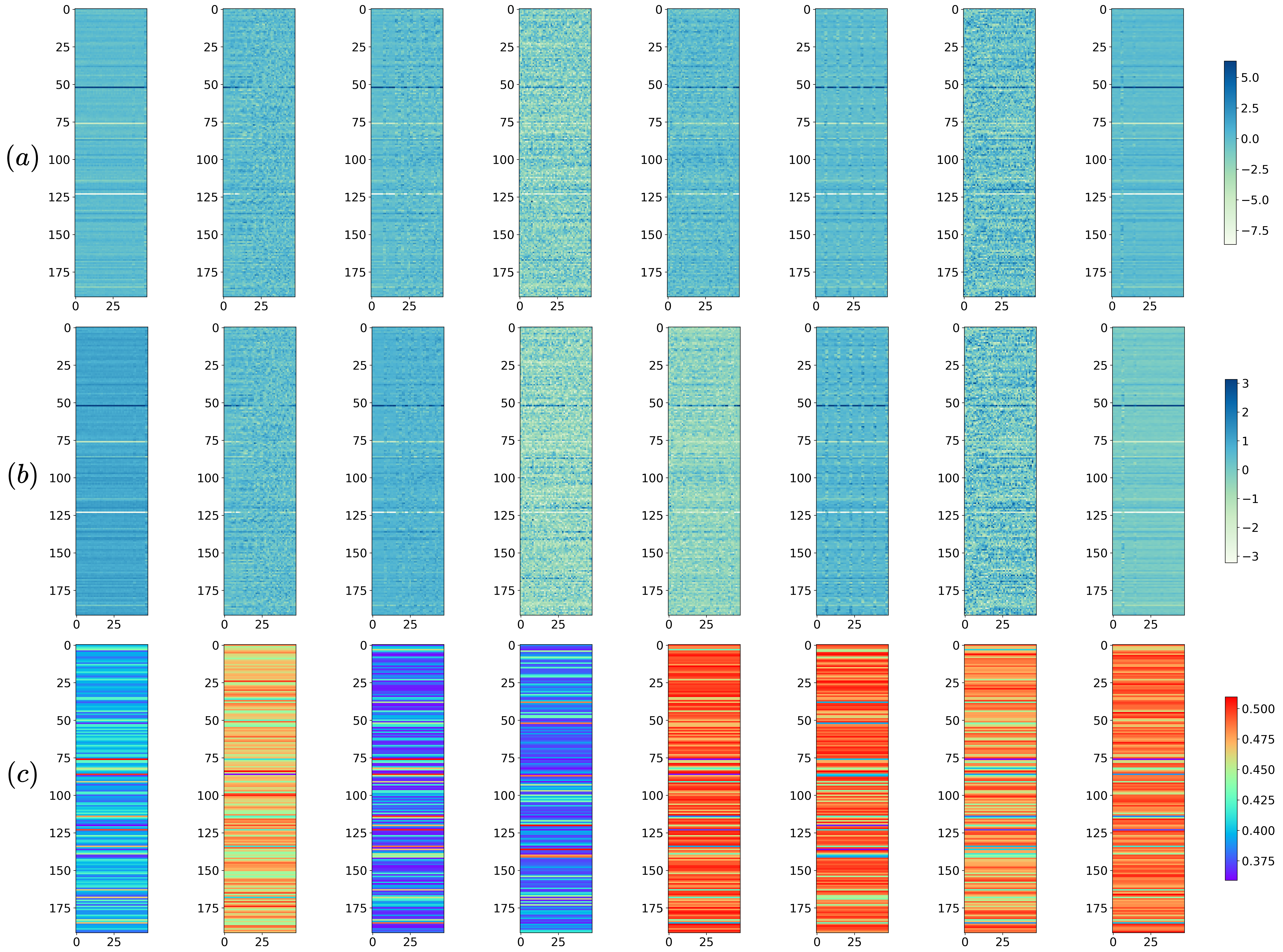}
        \caption{Eight example slices of the window sequence input into and output from the first WWA block at the 1/8 resolution stage. Each image denotes the window with 192 channels (horizontal axis) and 48 elements (vertical axis). Row $(a)$ represents the window slices before inputting them into the WWA block. Row $(b)$ represents the output from the WWA block, and $(c)$ denotes the quotients of the input and output windows that indicate the degree of difference.}
    \label{fig:attn_maps_1_8}
\end{figure*}

To assess the role of WWA in building window relevance at the global range, we visualize windowed features at the 1/8 resolution stage. The visualization results are shown in Fig. \ref{fig:attn_maps_1_8}. Combining the description of the example in Section \ref{WWA Example}, Fig. \ref{fig:attn_maps_1_8} demonstrates that the windowed feature representations output from WWA significantly differ from the input representations. By observing the quotients of the output and input feature representations, it is apparent that WWA assigns a weight to each channel of every window. This weight assignment indicates that WWA is capable of automatically associating and building the global interaction of a window sequence based on the feature representation of each window.

\section{Conclusion}

In this paper, we propose an unsupervised deformable image registration model based on the Swin transformer, dubbed as RFR-WWANet. RFR-WWANet exploits long-range spatial correlations to enhance feature representations. The restoring branch in RFR-WWANet can restore the resolution of feature maps from the Swin transformer block to a higher resolution stage to improve the ability of deep feature expression of the model and thus improve the contribution of Swin transformers in the model. The proposed WWA enhances the ability to build interaction between windows in a global range. Qualitative and quantitative evaluation results demonstrate that RFR-WWANet facilitates semantically meaningful correspondence of anatomical structures and provides state-of-the-art registration performance. Furthermore, ablation studies demonstrate the impact of the recovery branch and WWA on model performance, indicating the effectiveness and importance of the scheme of restoring feature resolution and WWA.

\section{Acknowledgements}
This work was supported by the National Nature Science Foundation of China [grant number 61772226]; Science and Technology Development Program of Jilin Province [grant number 20210204133YY]; The Natural Science Foundation of Jilin Province (Grant number No. 20200201159JC); Key Laboratory for Symbol Computation and Knowledge Engineering of the National Education Ministry of China, Jilin University.




\bibliographystyle{elsarticle-num-names}
\bibliography{egbib}





\end{document}